\documentclass{article}

\PassOptionsToPackage{numbers, compress}{natbib}

\usepackage[preprint]{neurips_2024}

\usepackage[utf8]{inputenc} %
\usepackage[T1]{fontenc}    %
\usepackage{hyperref}       %
\usepackage{url}            %
\usepackage{booktabs}       %
\usepackage{amsfonts}       %
\usepackage{nicefrac}       %
\usepackage{microtype}      %
\usepackage[table]{xcolor}         %

\usepackage{amsmath}
\usepackage{amssymb}
\usepackage{xspace}
\usepackage[]{xcolor}

\usepackage{graphicx}
\usepackage{booktabs}

\usepackage{wrapfig}
\usepackage{setspace}

\usepackage{algorithm}
\usepackage[noend]{algpseudocode}

\usepackage{soul}

\usepackage{enumitem}
\usepackage{mathrsfs}

\usepackage{amsmath}
\usepackage{amssymb}  %
\usepackage{graphicx} %

\usepackage{siunitx} %

\usepackage[htt]{hyphenat}
\usepackage{tcolorbox}

\makeatletter %

\newcommand*\sNeg[2][0mu]{\Neginternal{#1}{\snegslash}{#2}}

\newcommand*\Neginternal[3]{\mathpalette\Neg@{{#1}{#2}{#3}}}
\newcommand*\Neg@[2]{\Neg@@{#1}#2}
\newcommand*\Neg@@[4]{%
  \mathrel{\ooalign{%
    $\m@th#1#4$\cr
    \hidewidth$\m@th#3{#1}\mkern\muexpr#2*2$\hidewidth\cr
  }}%
}

\newcommand*\negslash[1]{\m@th#1\not\mathrel{\phantom{=}}}
\newcommand*\snegslash[1]{\rotatebox[origin=c]{60}{$\m@th#1-$}}
\newcommand*\ssnegslash[1]{\rotatebox[origin=c]{60}{$\m@th#1{\dabar@}\mkern-7mu{\dabar@}$}}
\newcommand*\sssnegslash[1]{\rotatebox[origin=c]{60}{$\m@th#1\dabar@$}}
\makeatother

\DeclareMathOperator*{\argmin}{\arg\!\min}

\newcommand\showhl{1}

\newcommand{\chl}[1]{%
    \ifnum 1=\showhl \relax
        {\color{cyan}#1}\else #1%
    \fi
}

\newcommand{\fchl}[1]{%
    \ifnum 1=\showhl \relax
        {\color{cyan}#1}\else #1%
    \fi
}

\definecolor{promptbg}{RGB}{234,239,246}
\definecolor{promptfrm}{RGB}{0,90,224}
\newtcolorbox{promptbox}[1]{colback=promptbg,colframe=promptfrm,size=small,fontupper=\scriptsize,fonttitle=\bfseries,title={#1}}

\title{Large Language Models are Effective Priors for Causal Graph Discovery}

\author{%
  Victor-Alexandru Darvariu$^{1}$,\enspace Stephen Hailes$^1$,\enspace Mirco Musolesi$^{1,2}$ \\
  $^1$University College London\quad $^2$University of Bologna \\
  \texttt{\{v.darvariu, s.hailes, m.musolesi\}@cs.ucl.ac.uk} \\
}

\begin{document}

\maketitle

\begin{abstract}

Causal structure discovery from observations can be improved by integrating background knowledge provided by an expert to reduce the hypothesis space. Recently, Large Language Models (LLMs) have begun to be considered as sources of prior information given the low cost of querying them relative to a human expert. 
In this work, firstly, we propose a set of metrics for assessing LLM judgments for causal graph discovery independently of the downstream algorithm. Secondly, we systematically study a set of prompting designs that allows the model to specify priors about the structure of the causal graph.
Finally, we present a general methodology for the integration of LLM priors in graph discovery algorithms, finding that they help improve performance on common-sense benchmarks and especially when used for assessing edge directionality.
Our work highlights the potential as well as the shortcomings of the use of LLMs in this problem space.

\end{abstract}

\section{Introduction}

The problem of causal discovery involves determining a probabilistic graphical model that establishes causal relationships among a set of random variables. This task holds fundamental importance in the sciences, as the resulting model can be utilized to answer observational, interventional, and counterfactual cause-and-effect queries~\citep{pearl2009causality,morgan2015counterfactuals,peters2017elements}. Causal discovery often operates within a constrained data environment where the acquisition of additional samples may be impractical, impossible, or ethically questionable. Consequently, numerous studies have explored the integration of appropriate prior knowledge, such as insights from human experts, to guide or bias the exploration of the vast hypothesis space. Prominent methods of incorporating prior knowledge involve the imposition of \textit{hard} constraints. These constraints may dictate the presence of specific edges~\citep{meek1995causal}, establish orderings among variables~\citep{cooper1992bayesian}, enforce ancestral relationships~\citep{chen2016learning}, or apply typing assumptions~\citep{brouillard2022typing}.

Large Language Models (LLMs) have recently been explored as sources of hard background knowledge, particularly when a human expert may be costly or unavailable~\citep{long2023causal,ban2023query,vashishtha2024causal}. This builds on earlier findings that suggest LLMs contain valuable information for causal reasoning~\citep{khetan2022causal,willig2022can,kiciman2023causal}. However, foundation models are rarely universally accurate, often failing to achieve 100\% accuracy even on simple arithmetic tasks~\citep{henighan2020scaling}. We therefore consider that the integration of LLM-derived knowledge should be \textit{soft} to prevent error propagation, an area that has received substantially less attention in prior work. Consequently, in this paper, we address the following questions:
\begin{enumerate}
	\item How can LLMs be judged on their abilities to identify causal relationships beyond plain accuracy and independently of the downstream causal discovery technique?
	\item What are the design choices in constructing the LLM prompt that lead to consistent, statistically significant improvements in model outputs?
	\item How can LLMs be integrated in a causal discovery method as soft background knowledge and under what type of regime are they beneficial?
\end{enumerate}
\begin{figure}[t]
\centering
\includegraphics[width=\textwidth,trim={3.5mm 0 3.5mm 0},clip]{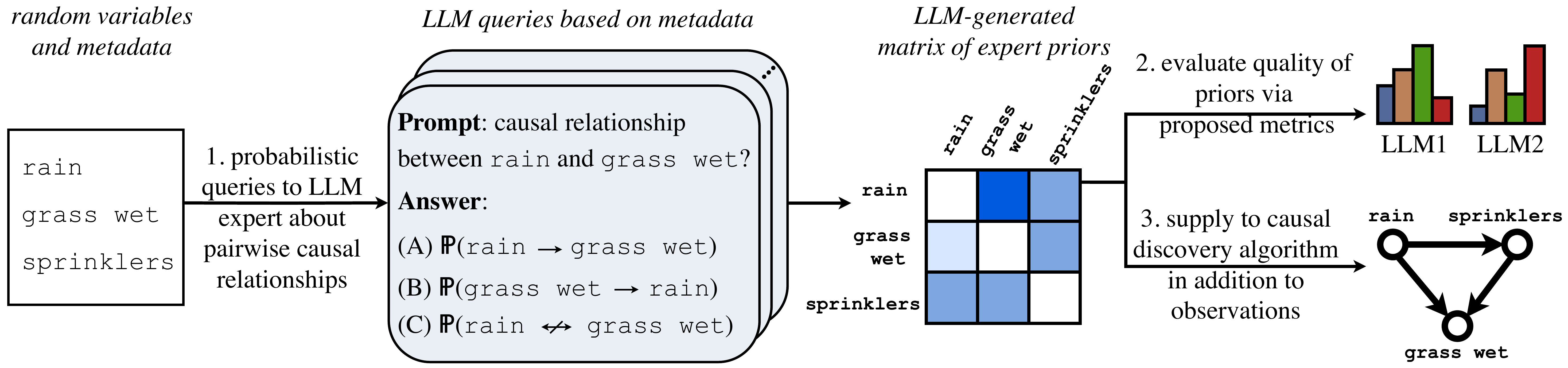} 
\caption{High-level summary of the methodology and contributions of the present work. Firstly, we formulate a probabilistic model of expert interaction for causal graph discovery and propose a set of metrics for assessing the quality of causal judgements supplied by LLMs. Secondly, we conduct an evaluation of several LLM architectures and prompt design choices using these metrics, showing that LLM background knowledge can convincingly outperform a null model on benchmarks requiring common-sense reasoning. Lastly, we integrate LLM-derived knowledge with a recent causal discovery method, finding that it is most beneficial for assessing the more likely direction of a relationship and in scenarios in which computational budgets are low.
}
\label{fig:cdllmillustration}
\end{figure}
Towards progress on these questions, we make the contributions listed below. They are summarized at a high level in Figure~\ref{fig:cdllmillustration}.
\begin{enumerate}
	\item We design a probabilistic expert interaction model for causal relationship extraction and a set of metrics for evaluating the judgments made by LLMs. Importantly, these metrics are designed to be independent of the specifics of the downstream causal discovery method. Our findings indicate that LLMs can convincingly exceed the performance level of random guessing, particularly when applied to benchmark datasets that do not necessitate specialized domain knowledge.
	\item We study the impact of $4$ prompt design choices on the proposed metrics in a case study with $3$ causal discovery datasets and $3$ open weight LLMs. We find that a 3-Way prompt (i.e., allowing the LLM to specify that a relationship does not exist) consistently and significantly improves metrics, while the impacts of other choices are situationally dependent;
	\item We present a methodology for the extraction and integration of LLM priors in graph discovery algorithms. In particular, we integrate LLM priors with a general causal discovery method~\citep{darvariu2023tree}.
We show that that the combination of LLM and mutual information priors for sampling edges can yield superior performance to baselines due to the ability of LLMs to judge the \textit{direction} of a relationship, especially in scenarios characterized by a low computational budget. 
The proposed methodology for extracting and integrating the priors is broadly applicable to other causal discovery algorithms that leverage pairwise edge scores.
\end{enumerate}

\section{Related Work}

\textbf{Hard background knowledge.} Many works have considered reducing the space of causal structures or biasing its navigation by leveraging \textit{background knowledge} (e.g., provided by an expert) alongside the available observations. Intuitive forms include imposing edges that are certainly present or absent~\citep{meek1995causal,decampos2007bayesian} or a given~\textit{ordering}, i.e., a sequence in which a node $X_i$ that precedes $X_j$ signifies there cannot exist an arc from $X_j$ to $X_i$~\citep{cooper1992bayesian}, which can reduce the search space and render it more regular~\citep{friedman2003being,teyssier2005ordering}. One can also specify \textit{ancestral constraints} of the form $X_i \rightsquigarrow X_j$, requiring the existence of a (possibly multi-hop) cause-and-effect relationship. Typing assumptions~\citep{brouillard2022typing} have also been considered recently, constraining the realizability of relationships based on expert-specified \textit{types}. This allows formally excluding those that cannot hold due to the nature of the random variable (e.g., the temperature of a city cannot possibly alter its altitude~\citep{peters2017elements}).

\textbf{Soft background knowledge.} A limitation of \textit{hard} knowledge is that the constraints derived from it may lead to the exclusion of the true solution in case it is wrong. \textit{Soft} background knowledge, on the other hand, makes certain structures more or less likely but does not impose formal restrictions.~\citep{mansinghka2006structured} considered soft priors based on a block model in which random variables belong to one of several classes that dictate the edge existence probability. Given that LLMs have been shown to successfully aid other methods heuristically while performing poorly in an autonomous mode for AI tasks such as planning~\citep{valmeekam2023planning}, their use for soft background knowledge is a promising yet under-explored direction.

\textbf{LLMs for causal reasoning.} Recently, LLM-derived priors have been investigated for various decision-making tasks~\citep{choi2022lmpriors}. Particularly in a low-data regime (as is typically the case in causal structure discovery), they have shown potential to act as a source of ``common-sense knowledge'' when metadata descriptions of the phenomena are available in natural language. An emerging body of work evidences that such foundational models contain useful information for causal reasoning tasks~\citep{khetan2022causal,willig2022can,kiciman2023causal,jin2024can}, such as distinguishing pairwise cause and effect~\citep{mooij2016distinguishing,choi2022lmpriors} as well as the full causal discovery problem in which a complex probabilistic structure must be identified.

\textbf{LLMs for causal structure discovery.}~\citet{long2022can} queried LLMs about the existence of causal relationships in the medical domain and evaluated the accuracy of their output against ground truth graphs. This technique constructs causal graphs directly and cannot leverage a dataset of observations. %
Further work by~\citet{long2023causal} used expert knowledge to reduce the size of a Markov Equivalence Class (as output by a pre-existing method) such that the ground truth graph is contained in the reduced set with high probability. The authors used hard expert judgements and a single 2-Way prompt design.~\citet{ban2023query} used LLMs to extract a set of ancestral constraints that a downstream method must obey (alternatively allowing violations that conflict with the observed data).~\citet{vashishtha2024causal} designed a technique for extracting an ordering by querying LLMs using node triplets, which was subsequently used in a downstream causal discovery method.

\textbf{Use of prior knowledge in Monte Carlo tree search.} CD-UCT~\citep{darvariu2023tree} is a variant of the UCT~\citep{kocsis_bandit_2006} algorithm that can be used for planning in Markov Decision Processes. Prior knowledge can be integrated with it at various levels~\citep{browne_survey_2012} in order to increase its effectiveness for a particular domain. A mechanism for doing so is designing the simulation policy that is used to sample actions (i.e., graph edges in this context) outside of the search tree, which may yield substantially better solutions than the default uniform random sampling of actions. The simulation policy can generally be hand-engineered~\citep{gelly2006modification,darvariu2023planning} or learned from interactions~\citep{gelly_combining_2007,silver_alphago_2016,anthony_thinking_2017}.

\section{Methods}\label{sec:methods}

\subsection{Causal Graph Discovery}

Let $G=(\mathcal{V},\mathcal{E})$ denote a Directed Acyclic Graph (DAG) with $d$ nodes and $m$ edges. Each node $v_i \in \mathcal{V}$ corresponds to a Random Variable (RV) $X_i$ that may be discrete or continuous. Edges $e_{i,j}$ between vertices $v_i$ and $v_j$ indicate a directional causal relationship between $X_i$ and $X_j$. Let $\mathbf{Pa}(X_i)$ denote the parent set of $X_i$ in the causal graph, i.e., RVs $X_k$ s.t. $e_{k,i} \ \in \mathcal{E}$. Variables $X_i$ are assumed to be independent of other RVs given their parent set: $\mathbb{P}\left(X_1, \ldots, X_d\right)=\prod_{i=1}^d \mathbb{P}\left(X_i \mid \mathbf{Pa}(X_i) \right)$.
Given a dataset $\mathbf{X} \in \mathbb{R}^{n \times d}$ of $n$ $d$-dimensional observations, the goal is to identify the true underlying DAG $G$. In \textit{score-based} methods, such as the CD-UCT model-based reinforcement learning approach we consider in our evaluation, this is formulated as optimizing a score function $f$ such as the widely used BIC~\citep{schwarz1978estimating}. Letting $\mathbb{D}^{(d)}$ denote the set of DAGs with $d$ nodes, the problem can be formalized as finding one of the graphs $G^*$ satisfying $G^*=\argmin_{G \in \mathbb{D}^{(d)}} {f(G)}$.

\subsection{Expert Interaction Model for Causal Relationship Extraction}\label{subsec:expertint}

We propose a probabilistic model of interactions with the expert for causal relationship extraction. Interactions are formalized via \textit{queries} $Q_{i,j}$ regarding the likelihood of a causal relationship existing between all possible pairs of RVs $(X_i,X_j)$. We let $\mathcal{Q}$ denote the set of all queries. Each query $Q_{i,j}$ is itself a random variable with 3 possible outcomes:

\begin{enumerate}
	\item Outcome $a$: the edge $e_{i,j}$ exists (i.e., there is a direct causal relationship from $X_i$ to $X_j$);
	\item Outcome $b$: the edge $e_{j,i}$ exists (i.e., there is a direct causal relationship from $X_j$ to $X_i$);
	\item Outcome $c$: neither $e_{i,j}$ nor $e_{j,i}$ exists in the causal graph (i.e., there is no direct causal relationship between $X_i$ and $X_j$ in either direction).
\end{enumerate}

In general, the expert provides probabilities $\mathbb{P}(Q_{i,j}=a)$, $\mathbb{P}(Q_{i,j}=b)$, $\mathbb{P}(Q_{i,j}=c)$ respectively for each query. We use $P_{i,j,a}, P_{i,j,b}, P_{i,j,c}$ as shorthands for these probabilities. Given outcomes $a,b,c$ are exhaustive and mutually exclusive, we have that $P_{i,j,a}+P_{i,j,b}+P_{i,j,c} = 1$. Note that the quantities $P_{i,j,a}$ and $P_{j,i,b}$ denote the probability of the same event, i.e., the existence of edge $e_{i,j}$. With a logically coherent expert, one would expect $P_{i,j,a}$ and $P_{j,i,b}$ to be equal. However, this tends to not be the case with LLMs, as our results will later show. We therefore issue queries in both directions and let $P_{i \rightarrow j}$ denote the arithmetic mean $(P_{i,j,a} + P_{j,i,b}) / 2$. Similarly, we let $P_{j \rightarrow i} = (P_{j,i,a} + P_{i,j,b}) / 2$ and $P_{i\sNeg{\leftrightarrow}j} = (P_{i,j,c} + P_{j,i,c}) / 2$. It is the case that $P_{i \rightarrow j} + P_{j \rightarrow i} + P_{i \sNeg{\leftrightarrow} j} = 1$.

The priors can be of different types. For this reason, we instantiate the general model, considering different types of priors forming a matrix $\mathbf{P}^{type}$ in which the entry at index $i,j$ is the value $P_{i \rightarrow j}^{type}$. 
For example, Uniform Random (UR) priors, which are equivalent to not using an expert, are denoted $\mathbf{P}^\text{UR}$ and a specific value as $P^\text{UR}_{i \rightarrow j}$. %

\subsection{Metrics for Evaluating Priors}\label{subsec:metrics}

The assessment of expert judgments as answers to a set of queries $\mathcal{Q}$ is a key aspect of the proposed approach. In fact, metrics for classification (such as those adopted in the existing literature discussed in the previous Section) are arguably insufficient since they do not fully capture the implications of using the expert knowledge as priors for causal discovery. For example, for an edge $e_{i,j}$ that exists in the true causal graph, it is not only the case that $P_{i \rightarrow j} > P_{j \rightarrow i}$ is desirable; we would also like the ratio between $P_{i,j}$ and $P_{j,i}$ to be as high as possible, so that the edge is substantially more likely to be sampled in the correct direction. We therefore propose a set of metrics that are better-suited for this setting and can be assessed independently of the downstream causal discovery method. Recall that $\mathcal{E}$ denotes the set of edges. It is convenient to also define the set of reverse edges $\mathcal{R} = \{ e_{j,i}\ |\ e_{i,j} \in \mathcal{E} \}$; and the set of non-edges $\mathcal{N} = \{ e_{i,j}\ |\ e_{i,j} \notin \mathcal{E} \land e_{j,i} \notin \mathcal{E} \}$. We then define the metrics as follows.

\textbf{Fraction of Correct Orientations (FCO)}. This metric captures the proportion of the true causal relationships that are judged as more likely in the correct direction. Letting $|\cdot|$ denote set cardinality,
$
\text{FCO}(\mathcal{Q}) = \frac{|\{ e_{i,j} \in \mathcal{E}\ | \ P_{i \rightarrow j} > P_{j \rightarrow i}\}|}{m}.
$

\textbf{True Edge to Reverse Edge (TERE)}. It captures the aforementioned ratio between the priors of the true edges and their reverse. Formally,
$
\text{TERE}(\mathcal{Q}) = \sum_{e_{i,j} \in \mathcal{E}} {\frac{P_{i \rightarrow j}}{P_{j \rightarrow i}} }.
$

\textbf{True Edge to Negative Edge (TENE)}. It captures the ratio between the probabilities of the true edges and non-edges, corresponding to the desirable characteristic that edges that exist in the ground truth graph are more likely to be sampled:
$
\text{TENE}(\mathcal{Q}) = \frac{\sum_{e_{i,j} \in \mathcal{E} }{P_{i \rightarrow j}}}{\sum_{e_{k,l} \in \mathcal{N} }{P_{k \rightarrow l}}}.
$

\textbf{Level Of Disagreement (LOD)}. Lastly, this quantity measures the lack of coherence in the expert judgments, with a value of $0$ indicating a coherent expert with respect to the directionality of the queries. It is defined as 
$
\text{LOD}(\mathcal{Q}) = \frac{\sum_{e_{i,j} \in \mathcal{E} \cup \mathcal{R} \cup \mathcal{N}}{(\text{abs}(P_{i,j,A} - P_{j,i,B}) + \text{abs}( P_{j,i,A} - P_{i,j,B}))}}{2d(d-1)}.
$

\subsection{Integrating Expert Priors with Causal Graph Discovery Algorithms}\label{subsec:integ}

Ultimately, the goal of this work is to leverage the priors in the causal discovery process. 
It is worth noting that the proposed approach can be integrated with any causal discovery algorithm that uses a matrix of pairwise edge scores.

In our evaluation, we combine the priors with the CD-UCT method~\citep{darvariu2023tree}, which is a general model-based reinforcement learning algorithm that optimizes a given score function. We choose this method for its flexibility, as it is applicable across many types of random variables and score functions, as well as the straightforward compatibility of the expert interaction model with this technique\footnote{
	To be more precise, using the priors together with CD-UCT biases the search while still allowing the maximization of the score function based on the observations. This does not exclude any of the valid causal graphs and realizes the requirement that the integration of background knowledge should be \textit{soft}, given the inherent characteristics of LLMs. The considered model of expert interactions, which yields a pairwise matrix of scores, can also be combined with other techniques in future work. For example, a pairwise matrix of scores is also used in LiNGAM~\citep{shimizu2006linear} and NOTEARS~\citep{zheng2018dags}, from which a DAG is then constructed by thresholding.
}. 
CD-UCT follows a simulation policy $\pi_\text{sim}$ for sampling valid edge choices from the action space $\mathcal{A}$ (which is defined in such a way to exclude edges that already exist or whose introduction would cause cycles). CD-UCT, much like the standard UCT method, uses a UR policy for performing this sampling -- and is thus equivalent to the null priors $\mathbf{P}^{\text{UR}}$ as defined above. Instead, using the expert information, the probability that the simulation policy $\pi_\text{sim}$ selects edge $e_{i,j}$ is defined as $\pi_\text{sim}(e_{i,j})=\frac{\exp (P_{i \rightarrow j} / \tau)}{\sum_{{e_{k,l}} \in \mathcal{A}} \exp \left( P_{k \rightarrow l} / \tau\right)}$, where the parameter $\tau$ controls the level of trust in the prior. When $\tau \to 0$ the valid edge with the largest prior value is chosen deterministically, while $\tau \to \infty$ approaches UR sampling.

\subsection{Mutual Information as a Baseline Prior}

As baseline, we consider a prior that quantifies the strength of association between two RVs. Mutual Information (MI) is an information-theoretic measure of pairwise dependence between two RVs $X_i$ and $X_j$. It represents the amount of information (e.g., measured in a unit such as bits) that knowing the value of $X_i$ reveals about $X_j$. For discrete variables, it can be computed as $P^{\text{MI}}_{i \rightarrow j} = I(X_i , X_j)=\sum_{x_i, x_j} \mathbb{P}_{X_i X_j}(x_i, x_j) \log \frac{\mathbb{P}_{X_i X_j}(x_i, x_j)}{\mathbb{P}_{X_i}(x_i) \mathbb{P}_{X_j}(x_j)}$.

In practice, the joint and marginal probabilities are calculated based on the observations $\mathbf{X}$. Techniques also exist for estimating MI for the continuous variable case~\citep{belghazi2018mutual}. Note that MI is a symmetric quantity: while it may signal the strength of a relationship, it does not indicate its directionality. %
In our approach, we propose combining MI and LLM-derived priors using the Hadamard product: $\mathbf{P}^{\text{MI}\odot\text{LLM}} = \mathbf{P}^\text{MI} \odot \mathbf{P}^\text{LLM}$. Doing so allows us to integrate an indication of the strength of a relationship with the ability of LLMs to provide information about its more likely directionality.
\section{Prompting Method and Design}\label{sec:prompting}

This Section describes the proposed method of interaction with the LLM-based expert. For illustration, we use the classic Bayesian network example with 3 RVs, in which \texttt{rain} and \texttt{sprinklers} both cause \texttt{grass wet}, and \texttt{rain} has a causal influence on \texttt{sprinklers}. Queries $\mathcal{Q}$ are issued to the LLM expert based on metadata $\mu_i$ for each RV $X_i$. The metadata is a textual description of what the RV measures, e.g., ``the presence of rain''. The LLM is presented with a prompt of the form:

\begin{promptbox}{}
\texttt{Among these three options which one is the most likely true: \\
(A) the presence of rain causes whether the grass is wet \\
(B) whether the grass is wet causes the presence of rain \\
(C) no causal relationship exists between the presence of rain and whether the grass is wet \\
The answer is: 
}
\end{promptbox}

Subsequently, it is asked to generate the next token, and the probabilities of the $\texttt{(A)}$, $\texttt{(B)}$, and $\texttt{(C)}$ tokens are measured. These correspond directly to the $\mathbb{P}(Q_{i,j}=a)$, $\mathbb{P}(Q_{i,j}=b)$, and $\mathbb{P}(Q_{i,j}=c)$ probabilities as defined in Section~\ref{subsec:expertint}. It may be the case that the token probabilities do not strictly sum to $1$ (i.e., other tokens have vanishingly small probabilities), in which case they are renormalized.

Let us now discuss several considerations regarding prompt design that will be assessed experimentally in the next Section. We refer to these choices as \textit{traits}.

\textbf{3-Way versus 2-Way}. The prompt shown above has three possible answers, and we therefore refer to it as \textit{3-Way}. The prompt used by~\citet{long2023causal}, in contrast, does not allow specifying that the most likely outcome is the absence of the relationship. We refer to this design as \textit{2-Way}.

\textbf{Variable List}. We consider, prior to the evaluation of a specific query, providing the LLM with a list of metadata descriptions of the RVs. The rationale is that, depending on all RVs to be considered, the validity of the relationship $X_i \to X_k$ is possibly conditional on an intermediary variable $X_j$ such that $X_i \to X_j$ and $X_j \to X_k$. The two cannot be distinguished without enumerating all RVs:
\begin{promptbox}{}
\texttt{A list of all the phenomena to be considered follows.\\
the presence of rain \\
whether the grass is wet \\
whether the sprinklers are switched on
}
\end{promptbox}
\textbf{Example}. We also consider providing an unrelated example of a question and answer pair such as the query shown above, so that the LLM is familiarized with the expected input-output format.

\textbf{Priming}. The final trait under consideration is whether the LLM is provided with a description of the domain and task prior to being queried, such as:
\begin{promptbox}{}
\texttt{A User interacts with an Expert. The Expert has profound knowledge and experience in the impact of meteorological phenomena. 
The Expert answers queries about possible causal relationships between two phenomena using their knowledge about cause and effect in this domain.
The Expert answers either (A) or (B) corresponding to the direction of the causal relationship, or (C) if no causal relationship exists.}
\end{promptbox}

The LLM is prompted with a query of a causal nature in either the 3-Way or 2-Way format, optionally preceded by a component corresponding to the Variable List, Example, or Priming traits. The LLM is prompted separately for each individual query $Q$ (i.e., the context is reset). More details, including full prompt templates, are provided in the Appendix. 

\section{Experiments}\label{sec:experiments}

\subsection{Experimental Setup}\label{subsec:expsetup}

\textbf{Datasets.} We use the classic benchmark datasets in the Bayesian networks literature, originally part of the Bayesian Network Repository~\citep{bnrepository}: \textit{Asia}~\citep{lauritzen1988local} ($d=8, m=8$), \textit{Child}~\citep{spiegelhalter1992learning} ($d=20, m=25$), and \textit{Insurance}~\citep{binder1997adaptive} ($d=27, m=52$). We use $n=1000$ samples for each. The metadata descriptions of the RVs are adapted from~\citep{long2023causal}. Further details are supplied in the Appendix.

\textbf{Choice of LLMs.} We make use of open-weights LLMs to conduct our evaluation, which renders our results fully reproducible. Specifically, for the primary experiments, we utilize the LLaMA2-7B~\citep{touvron2023llama2}, LLaMA3-8B~\citep{meta2024llama3}, and Mistral-7B~\citep{jiang2023mistral} models. We work with models of this size as they have shown good performance on commonsense reasoning tasks~\citep{touvron2023llama2} while being sufficiently fast when performing inference, an important consideration given the scale of our evaluation. In fact, we consider all possible pairwise relationships for several datasets, models, and prompt designs repeated across random seeds. To assess the possible impacts of larger models, we also carry out the evaluation with the LLaMA2-13B and LLaMA2-70B models (on the \textit{Asia} dataset only due to computational budget limitations) and report the results in the Appendix. We leverage the fine-tuned ``Chat" or ``Instruction'' variants. Further technical details are provided in the Appendix.

\textbf{Output stochasticity.} When performing LLM inference, we use a temperature of $0$, resulting in the model outputting the highest-probability token. This choice is made so that outputs are as ``factual'' as possible. To obtain statistically meaningful results, we introduce stochasticity by varying the \textit{verb} used when prompting out of $20$ choices including \texttt{causes}. The full list is given in the Appendix. 

\textbf{Evaluation methodology.} We report results aggregated over $200$ random seeds and display 95\% confidence intervals where relevant\footnote{In a future version, our implementation, data, and instructions will be publicly released in order to ensure full reproducibility of all the reported results.}. We repeat experiments with priors produced with different causal verbs; more specifically, each of the $20$ prior matrices $\mathbf{P}$ obtained with different verbs is used with $10$ random initializations of the causal discovery procedure. For the standalone LLM evaluations, we report the metrics proposed in Section~\ref{subsec:metrics}. For the full causal discovery problem, we report the Structural Hamming Distance (SHD), i.e., the minimum number of edge additions, deletions, and reversals for transforming the output graph into the ground truth causal graph.

\subsection{Standalone LLM Evaluation}~\label{subsec:standaloneresults}

\begin{figure}[t]
\centering
\includegraphics[width=\textwidth]{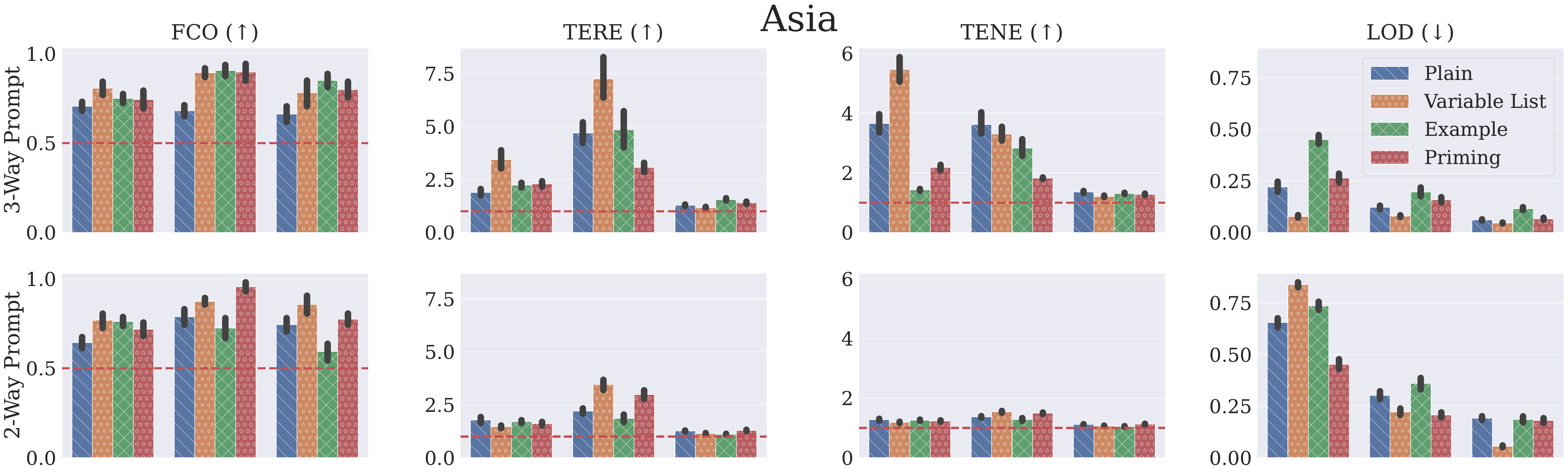} \\
\includegraphics[width=\textwidth]{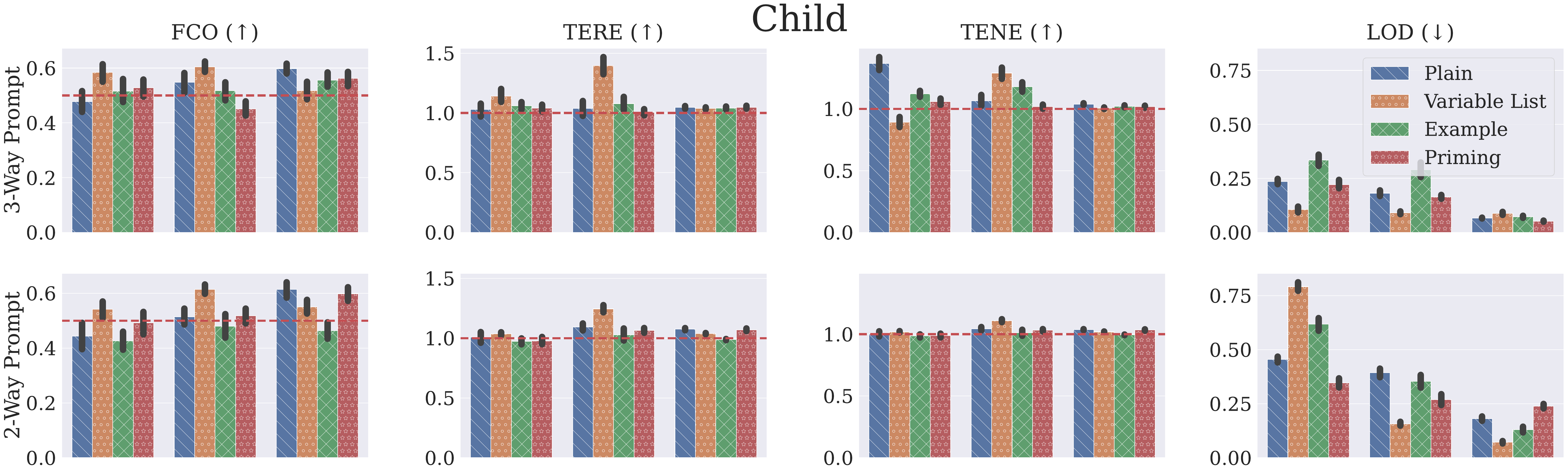} \\
\includegraphics[width=\textwidth]{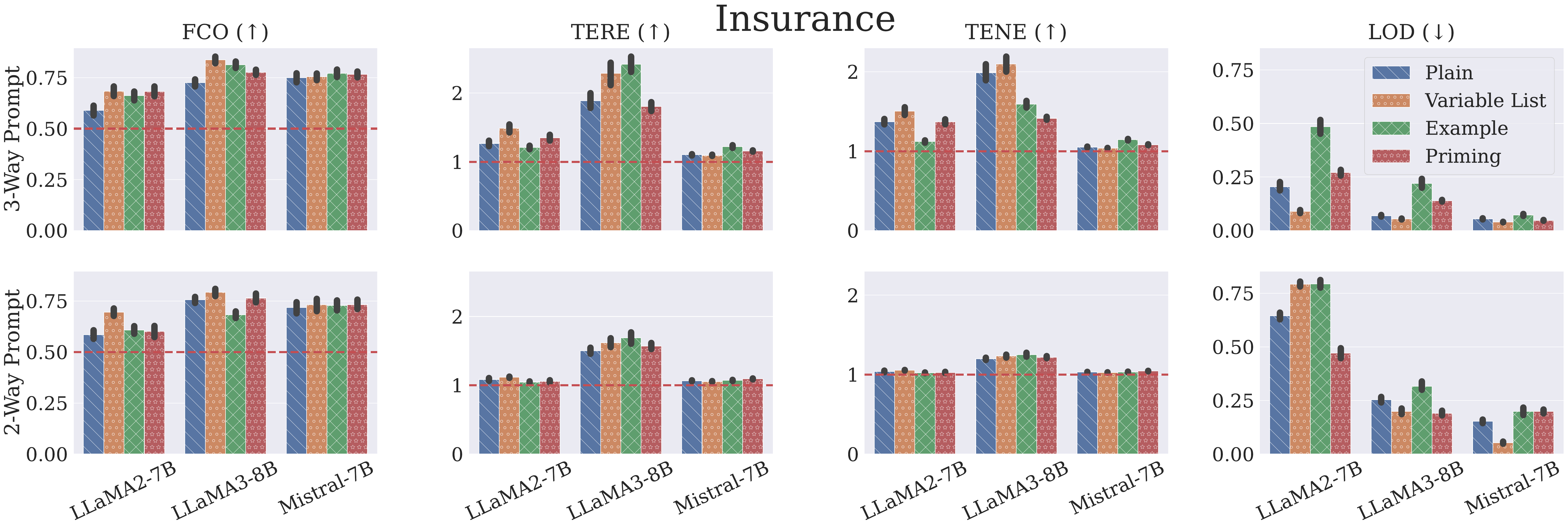} 
\caption{Results for the metrics defined over probabilistic LLM judgments. The red lines indicate the values that would be obtained by UR priors. In most cases, LLMs show better results than those obtained with UR, and hence can serve as useful priors for causal discovery. LLM performance is weaker on the \textit{Child} dataset, which requires specialist domain knowledge. LLaMA models yield better TERE and TENE values at the expense of a higher LOD.}
\label{fig:metrics}
\end{figure}

Figure~\ref{fig:metrics} shows the results obtained by measuring the proposed metrics on the probabilistic LLM outputs. Each column displays the value of a particular metric and the rows correspond to results for 3-Way and 2-Way prompting respectively for each of the 3 datasets. Within each plot, the bars are split by the specific LLM and prompt design. The horizontal red line, where applicable, corresponds to the use of UR priors. Overall, the LLMs typically surpass this threshold, suggesting their use as priors for causal discovery would surpass UR priors. The results are poorer on the \textit{Child} dataset ($d=20$), which arguably requires more specialized knowledge -- e.g., an edge exists between the RVs whose descriptions are \texttt{hypoxia when breathing oxygen } and \texttt{ level of oxygen in the right up quadricep muscle}, which may not be straightforward for an untrained human expert. Results for the \textit{Insurance} benchmark, which requires more ``common sense'', are substantially better despite the larger graph size ($d=27$). We complement these results with statistical tests that can be found in Table~\ref{tab:ttests} in the Appendix.

The dataset under consideration and prompt design are strong determinants of performance and there is no universal winner. Nevertheless, LLaMA3-8B generally outperforms LLaMA2-7B in all metrics. Mistral-7B also obtains better edge orientations (FCO) than LLaMA2-7B in two thirds of cases. The LLaMA models exhibit higher disagreement (LOD) and hence provide more inconsistent answers. This is also reflected in the higher TERE and TENE obtained by the LLaMA models. Regarding the different prompt traits, 3-Way prompting improves FCO, TERE and TENE, while reducing LOD in all settings tested.
This specific result also confirms a finding of a study of LLM prompting based on one LLM and one dataset presented in~\citep{kiciman2023causal}. Hence, this query structure is preferable when eliciting probabilistic background knowledge from LLMs. Furthermore, the Variable List trait consistently improves metrics for the LLaMA models, with inconclusive impacts on Mistral-7B. Finally, the other traits result in metric differences that, while generally statistically significant, do not show a consistent pattern across datasets and LLMs. 

\vspace{1cm}
\subsection{Priors for Causal Graph Discovery}~\label{subsec:cdresults}

\begin{table}[t]
\caption{SHD obtained when constructing causal graphs by greedily choosing the edges with the top-$m$ largest priors. LLM priors lead to causal graphs that are more accurate than those sampled uniformly at random. We observe that MI can be used as a basis for a strong prior that outperforms LLMs when considered in isolation. However, the best-performing prior is obtained by combining MI and LLM priors, with the latter providing the more likely \textit{directionality} of the relationship.
}
\label{tab:greedy}
\begin{center}
\resizebox{\textwidth}{!}{
\begin{tabular}{l|S[table-format=2.3]S[table-format=2.3]S[table-format=2.3]S[table-format=2.3]S[table-format=2.3]S[table-format=2.3]S[table-format=2.3]S[table-format=2.3]}
\toprule
{Dataset} & \textsc{\small MI} &  \textsc{\small LLaMA2-7B} &  \textsc{\small LLaMA3-8B} &  \textsc{\small Mistral-7B} &  \textsc{\small MI$\odot$LLaMA2-7B} &  \textsc{\small MI$\odot$LLaMA3-8B} &  \textsc{\small MI$\odot$Mistral-7B} &  \textsc{\small Uniform Random} \\
\midrule
Asia & 7.020 &      7.470 &      8.370 &       9.295 &         6.595 &         $\mathbf{6.240}$ &          6.365 &          12.620 \\
Child & 25.715 &     46.930 &     43.115 &      46.505 &        31.080 &        28.015 &         $\mathbf{23.515}$ &          45.300 \\
Insurance & 68.470 &     83.830 &     79.550 &      84.615 &        63.385 &        $\mathbf{55.995}$ &         63.200 &          92.080 \\
\bottomrule
\end{tabular}

}
\end{center}
\end{table}

For the full causal discovery problem, we use priors derived with the 3-Way Plain prompting style for Mistral-7B and 3-Way Variable List for the LLaMA models, as informed by the results above. We first evaluate the priors' quality in a simplified setting by constructing the causal graph for the highest-scoring top-$m$ edges. This is performed incrementally starting with the highest-value edge, breaking ties arbitrarily, and excluding edges that would introduce cycles. Table~\ref{tab:greedy} shows the SHD to the true causal graph averaged over $200$ random seeds. We find that the standalone LLM priors, while better than UR priors, fare substantially worse than MI. The best-performing priors are those obtained by combining MI and LLMs, therefore leveraging the foundation models' abilities to judge the more likely directionality of the relationships.

\begin{figure}[t]
\centering
\includegraphics[width=\textwidth]{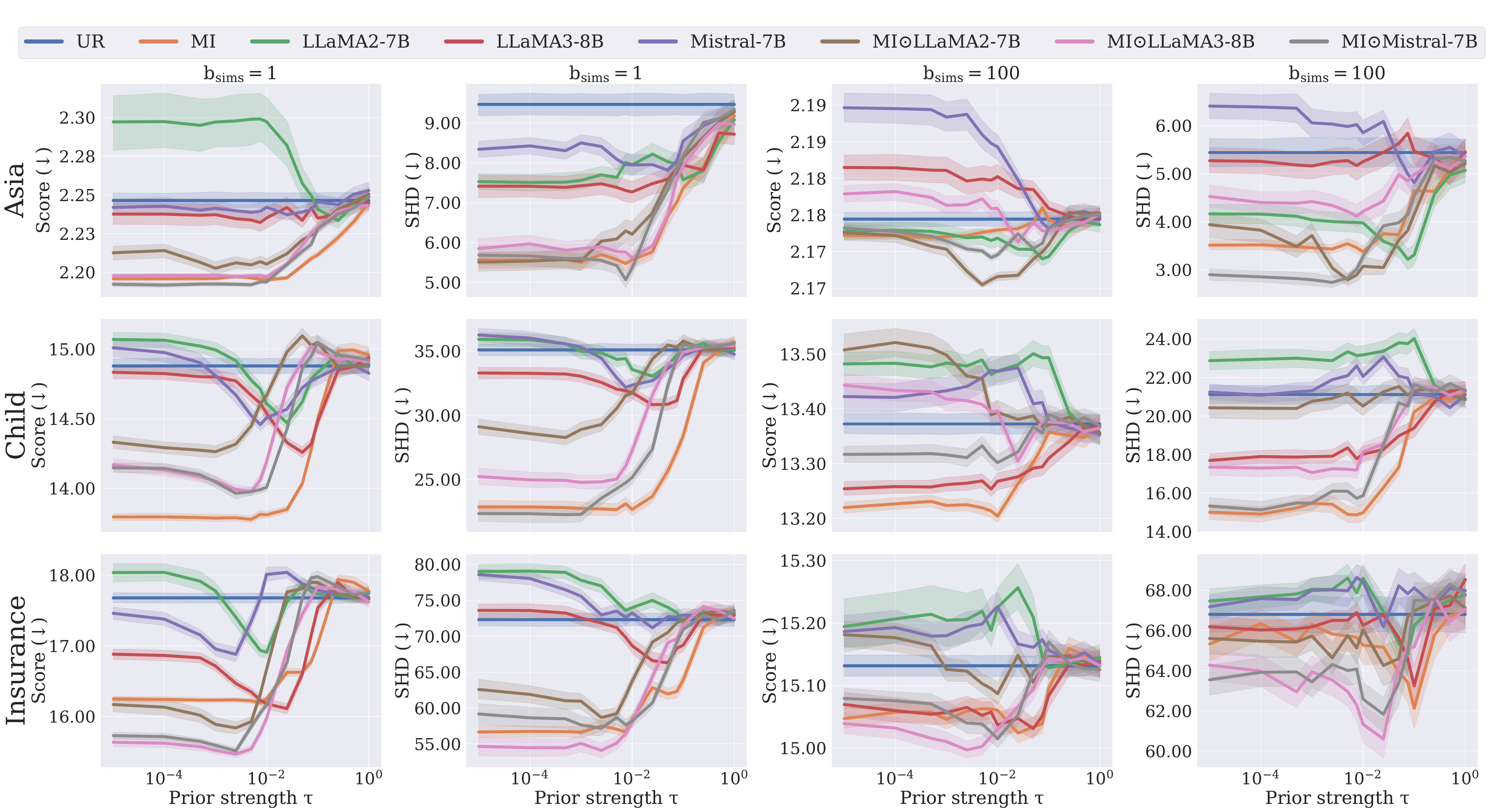} 
\caption{Results obtained using CD-UCT with various priors. The leftmost two and rightmost two columns show the results with computational budget multipliers of $1$ and $100$ respectively. MI and MI$\odot$LLM priors outperform the default UR priors over edges, while LLM priors by themselves do not always do so. The benefit of using priors is more pronounced with lower simulation budgets. An intermediate level of trust in the prior tends to result in the best outcome.
}
\label{fig:cductpriors}
\end{figure}

In Figure~\ref{fig:cductpriors}, we display the results obtained when equipping the CD-UCT causal discovery method with various priors. The first two and last two columns show results obtained with a low ($b_\text{sims}=1$) and high ($b_\text{sims}=100$) computational budget of simulations, respectively. The $x$-axis shows the prior strength $\tau$, while the $y$-axis measures the metric of interest (score function value of the output graph and SHD to the ground truth graph respectively).

We find that MI and MI$\odot$LLM priors consistently outperform the standard UR simulation policy of CD-UCT, while standalone LLM priors are not always beneficial. The gain when using the priors tends to be stronger with a low computational budget (i.e., when outputs have to be generated quickly), as indicated by the larger differences w.r.t. the blue line in the leftmost two columns. Regarding the level of trust $\tau$ in the priors, a middle ground value tends to perform best. As $\tau$ increases and priors matter less, metrics converge to those obtained by the standard CD-UCT with a UR simulation policy. Interestingly, in some cases such as LLaMA2-7B and the \textit{Asia} dataset, the LLM priors can lead to a more truthful causal graph (low SHD) despite high score function values. This indicates that the LLM judgments are accurate despite the relationship not being supported by the data. Table~\ref{tab:sdiff} shows the percentage decreases in SHD relative to CD-UCT with UR priors obtained with the $\tau$ value that leads to the best-scoring graphs, indicating that most considered priors attain improvements over UR. Combining MI and LLM priors performs best on \textit{Asia} and \textit{Insurance}, while MI by itself does best on the \textit{Child} benchmark on which the LLM judgments are poor in general.

\begin{table}[t]
\caption{Percentage decrease in SHD obtained by CD-UCT with the specified priors over the standard UR priors (higher is better). Generally, the considered priors improve on the standard method, indicated by cells containing values above $0$. Combining MI and LLM priors can lead to the best improvement over standard CD-UCT. As in the standalone evaluation, LLM performance is poorer on the highly specialized \textit{Child} dataset.
}{}
\label{tab:sdiff}
\begin{center}
\resizebox{\textwidth}{!}{
\begin{tabular}{ll|S[table-format=2.3]S[table-format=2.3]S[table-format=2.3]S[table-format=2.3]S[table-format=2.3]S[table-format=2.3]S[table-format=2.3]}
\toprule
Dataset  & $b_\text{sims}$ &  \textsc{\small MI} &  \textsc{\small LLaMA2-7B} &  \textsc{\small LLaMA3-8B} &  \textsc{\small Mistral-7B} &  \textsc{\small MI$\odot$LLaMA2-7B} &  \textsc{\small MI$\odot$LLaMA3-8B} &  \textsc{\small MI$\odot$Mistral-7B} \\
\midrule
     Asia &  $10^0$ & $\mathbf{39.903}$ &     13.102 &     19.599 &      11.803 &        39.199 &        38.062 &         38.603 \\
      &  $10^1$ & 33.187 &     44.274 &      1.678 &      -1.459 &        $\mathbf{46.244}$ &        32.677 &         38.877 \\
      & $10^2$ & 37.280 &     40.333 &      5.643 &       9.436 &        $\mathbf{48.844}$ &        26.272 &         45.051 \\
\midrule
    Child &   $10^0$ & $\mathbf{35.277}$ &      9.820 &     14.147 &       8.654 &        18.095 &        29.615 &         34.673 \\
     &  $10^1$ & $\mathbf{29.211}$ &     -1.089 &      5.902 &      -1.300 &        -1.669 &        21.342 &         22.958 \\
     & $10^2$ & $\mathbf{28.670}$ &     -0.191 &     14.681 &      -1.767 &         0.286 &        11.506 &         23.872 \\
\midrule
Insurance &   $10^0$ & 22.999 &     -0.095 &      9.310 &       0.530 &        20.748 &        $\mathbf{25.787}$ &         23.686 \\
 &  $10^0$ &  9.447 &      1.057 &      2.790 &      -0.050 &         3.336 &        $\mathbf{12.503}$ &          7.700 \\
 & $10^2$ &  2.389 &      0.052 &      1.940 &      -0.592 &         2.614 &         4.396 &          $\mathbf{5.580}$ \\
\bottomrule
\end{tabular}

}
\end{center}
\end{table}

\section{Conclusions}\label{sec:conclusion}

\textbf{Contributions.} In this work, we have examined the potential of using LLM priors for causal discovery. We have formulated a probabilistic model of expert interaction that allows for the soft integration of prior knowledge.
Furthermore, we have proposed a suite of metrics for evaluating LLM judgments separately from the downstream causal graph discovery method. %
Our evaluation shows that LLMs can be used to extract informative priors. 
We have also considered several design choices for prompts and found that 3-Way queries, which allow the model to specify that no direct causal relationship exists, consistently lead to more accurate causal graph reconstruction. 
Finally, we integrated LLM priors with a state-of-the-art causal discovery method using mutual information for sampling, demonstrating that combining them with mutual information priors can lead to superior performance. This is largely due to the ability of LLMs to determine the more likely direction of a relationship, which is particularly useful in scenarios with a limited computational budget.

\textbf{Limitations.} Regarding the limitations of this work, we acknowledge that while considering the $d*(d-1)$ pairwise relationships is feasible at this scale, larger networks would require a form of pre-processing to establish a restricted set of possible parents for each RV. This would reduce the number of queries and could be based on the collected observations or expert knowledge. Furthermore, LLMs may have been exposed to the considered benchmarks during training.

\textbf{Future work.} %
A more interactive model of expert interaction can be considered;
however, such a model would potentially require substantially more LLM inferences.
We also note that fine-tuning the LLM for causal reasoning has the potential to improve performance; however, this might require a significant amount of data for extracting the ``true'' causal graphs.

\newpage

\begin{ack}
This work was supported by the Turing’s Defence and Security programme through a partnership with the UK government in accordance with the framework agreement between GCHQ \& The Alan Turing Institute.
\end{ack}

{
\small

\bibliographystyle{plainnat}
\bibliography{bibliography}
}

\appendix

\section{Experiment and Implementation Details}\label{sec:appendixdetails}

\textbf{Code and data.} In a future version, we will release source code accompanied by detailed instructions for setting up and reproducing the experiments presented in the main text. The Bayesian Network Repository~\citep{bnrepository} makes the datasets available without licensing restrictions.

\textbf{CD-UCT method and parameters.} We implement CD-UCT as described in~\citep{darvariu2023tree}. In the experiments, we use an exploration parameter $C_p=0.025$, search horizon $h=16$, and simulation budget parameters $b_\text{sims} \in \{10^0, 10^1, 10^2\}$. We consider $15$ possible values for the prior strength $\tau \in \{0.00001, 0.0001, 0.0005, 0.001, 0.0025, 0.005, 0.0075, 0.01, 0.025, 0.05, 0.075, 0.1, 0.25, 0.5, 1\}$. The other parameters are held constant so that the different types of priors can be compared equitably. 

\textbf{LLM inference and quantization.} To perform inference, we make use of the \href{https://github.com/ggerganov/llama.cpp}{\texttt{llama.cpp}} library, whose custom binary format is compatible with a variety of models and hardware architectures, allowing LLM inference on CPUs. We use 6-bit (i.e., \texttt{Q6\_K}) post-training quantization (PTQ) for all models except for the larger LLaMA2-70B, for which we use the 5-bit \texttt{Q5\_K\_M} quantization. This substantially reduces hardware requirements with fairly minimal quality loss (i.e., the specified 6-bit quantization yields a perplexity 0.1\% higher compared to the unquantized 16-bit original model, as shown through \href{https://github.com/ggerganov/llama.cpp/pull/1684}{benchmarks performed by the library authors}).

\textbf{Computational infrastructure.} Experiments were conducted exclusively on commodity CPUs on an in-house High Performance Computing (HPC) cluster. On this infrastructure, the standalone LLM experiments (Section~\ref{subsec:standaloneresults}) and the full causal discovery experiments (Section~\ref{subsec:cdresults}) took approximately $718$ and $616$ days of single-core CPU time respectively.

\section{Additional Results}

\textbf{Statistical tests for prompt traits.} In Table~\ref{tab:ttests}, we present the results of statistical tests to examine the impacts of the different prompt traits, complementing the results in Section~\ref{subsec:standaloneresults}. Each cell displays the difference in the metric obtained when the specified trait is present versus absent. We perform a paired t-test and highlight statistically insignificant values at a 95\% confidence level in light gray. Approximately $80\%$ ($115/144$) of the comparisons yield statistically significant results and support the interpretation given in the main text.

\begin{table}[h]
\caption{Mean differences obtained \textit{with} and \textit{without} the specified prompt trait. We find that 3-Way prompting improves the values of all the metrics across the board, and should be preferred when querying LLMs about causal relationships. The other prompt design choices do not yield universally better metric values, resulting in effects that are dependent on the dataset and specific model.}
\label{tab:ttests}
\begin{center}
\resizebox{0.65\textwidth}{!}{
\begin{tabular}{lll|rrrr}
\toprule
Trait & Dataset & Model & FCO (↑) & TERE (↑) & TENE (↑) & LOD (↓) \\
\midrule
3-Way Prompt & Asia & LLaMA2-7B & 0.0281 & 0.8259 & 1.9447 & -0.4173 \\
 & & LLaMA3-8B & \cellcolor{gray!30}0.0087 & 2.3495 & 1.4726 & -0.1338 \\
 &  & Mistral-7B & \cellcolor{gray!30}0.0312 & \cellcolor{gray!30}0.1579 & 0.1998 & -0.0804 \\
 & Child & LLaMA2-7B & 0.0505 & 0.0685 & 0.1121 & -0.3281 \\
 & & LLaMA3-8B & \cellcolor{gray!30}-0.0012 & \cellcolor{gray!30}0.0219 & 0.0902 & -0.1109 \\
 &  & Mistral-7B & 0.0020 & 0.0004 & 0.0002 & -0.0850 \\
 & Insurance & LLaMA2-7B & \cellcolor{gray!30}0.0320 & 0.2530 & 0.3061 & -0.4128 \\
 & & LLaMA3-8B & 0.0384 & 0.5052 & 0.5467 & -0.1186 \\
 &  & Mistral-7B & \cellcolor{gray!30}0.0322 & 0.0737 & 0.0449 & -0.0970 \\
\midrule
Variable List & Asia & LLaMA2-7B & 0.1125 & 0.6127 & 0.8624 & \cellcolor{gray!30}0.0201 \\
 & & LLaMA3-8B & \cellcolor{gray!30}0.1500 & \cellcolor{gray!30}1.9033 & -0.0683 & -0.0604 \\
 &  & Mistral-7B & 0.1156 & -0.1176 & -0.0965 & -0.0759 \\
 & Child & LLaMA2-7B & 0.1020 & 0.0698 & -0.2295 & 0.1039 \\
 &  & LLaMA3-8B & 0.0780 & 0.2564 & 0.1449 & -0.1637 \\
 &  & Mistral-7B & \cellcolor{gray!30}-0.0720 & \cellcolor{gray!30}-0.0224 & -0.0254 & -0.0438 \\
 & Insurance & LLaMA2-7B & 0.1024 & 0.1258 & 0.0734 & 0.0164 \\
 &  & LLaMA3-8B & \cellcolor{gray!30}0.0745 & 0.2581 & 0.0731 & -0.0350 \\
 &  & Mistral-7B & 0.0096 & -0.0092 & -0.0095 & -0.0572 \\
\midrule
Example & Asia & LLaMA2-7B & 0.0813 & 0.1317 & -1.1198 & \cellcolor{gray!30}0.1556 \\
 &  & LLaMA3-8B & 0.0813 & \cellcolor{gray!30}-0.1025 & -0.4334 & 0.0666 \\
 &  & Mistral-7B & \cellcolor{gray!30}0.0187 & 0.0576 & -0.0569 & 0.0239 \\
 & Child & LLaMA2-7B & 0.0100 & -0.0021 & -0.1309 & 0.1314 \\
 &  & LLaMA3-8B & -0.0320 & \cellcolor{gray!30}-0.0116 & 0.0403 & 0.0342 \\
 &  & Mistral-7B & \cellcolor{gray!30}-0.0960 & \cellcolor{gray!30}-0.0458 & \cellcolor{gray!30}-0.0304 & -0.0229 \\
 & Insurance & LLaMA2-7B & \cellcolor{gray!30}0.0476 & -0.0482 & -0.1336 & 0.2147 \\
 &  & LLaMA3-8B & 0.0072 & \cellcolor{gray!30}0.3592 & \cellcolor{gray!30}-0.1709 & 0.1069 \\
 &  & Mistral-7B & \cellcolor{gray!30}0.0163 & 0.0617 & 0.0477 & 0.0325 \\
\midrule
Priming & Asia & LLaMA2-7B & 0.0563 & \cellcolor{gray!30}0.1134 & -0.7553 & -0.0797 \\
 &  & LLaMA3-8B & 0.1924 & -0.4295 & \cellcolor{gray!30}-0.8362 & -0.0284 \\
 &  & Mistral-7B & 0.0844 & 0.0732 & -0.0341 & -0.0032 \\
 & Child & LLaMA2-7B & 0.0490 & \cellcolor{gray!30}-0.0096 & -0.1615 & -0.0609 \\
 &  & LLaMA3-8B & -0.0463 & -0.0285 & -0.0295 & -0.0714 \\
 &  & Mistral-7B & -0.0260 & -0.0015 & -0.0114 & 0.0216 \\
 & Insurance & LLaMA2-7B & 0.0543 & \cellcolor{gray!30}0.0262 & \cellcolor{gray!30}-0.0072 & -0.0539 \\
 &  & LLaMA3-8B & 0.0297 & -0.0080 & -0.2757 & 0.0038 \\
 &  & Mistral-7B & 0.0154 & 0.0415 & 0.0213 & 0.0199 \\
\bottomrule
\end{tabular}

}
\end{center}
\end{table}

\begin{figure}[t]
\centering
\includegraphics[width=0.92\textwidth]{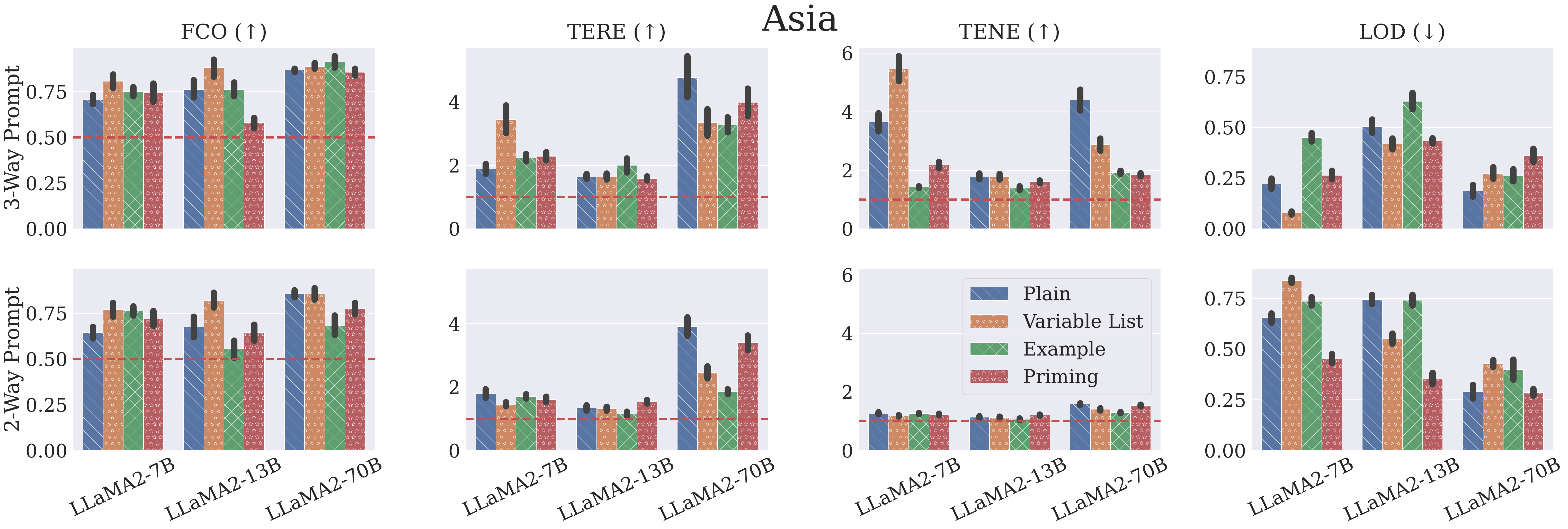} \\
\caption{Metrics results for larger LLaMA models on the \textit{Asia} dataset. While some improvements are exhibited, models with higher parameter counts do not lead to substantially better performance, echoing results on commonsense reasoning benchmarks.}
\label{fig:llamalargemetrics}
\end{figure}

\textbf{Results with larger LLaMA2 models.} We conduct the same experiments as reported in Sections~\ref{subsec:standaloneresults} and~\ref{subsec:cdresults} using the larger LLaMA2-13B and LLaMA2-70B models. Due to computational budget limitations, we perform this on the Asia dataset only. The obtained standalone metrics and causal discovery results are shown in Figure~\ref{fig:llamalargemetrics} and Table~\ref{tab:llamalarge} respectively. While larger models do present some improvements (e.g., the 70B variant yields an average of 84\% FCO compared to the 74\% FCO obtained by the 7B version), the differences are fairly small and not always consistent (e.g., the larger 13B variant obtains 71\% FCO on average). This reflects that the considered task resembles commonsense reasoning or world knowledge benchmarks more closely, for which differences between LLM sizes are less significant compared to language understanding or mathematical reasoning tasks~\citep{touvron2023llama2}. 
The differences are even less consistent when the priors are used together with the causal discovery algorithm. This is a scenario in which they interact with the maximization of the score function and the Mutual Information computed on the dataset.

\begin{table}[t]
\caption{SHD obtained on Asia by CD-UCT with priors derived from LLaMA2 variants of different sizes. The results with respect to model size are inconclusive, given that they are affected by the interaction with the score function maximization procedure in the causal discovery method.}
\label{tab:llamalarge}
\begin{center}
\resizebox{0.92\textwidth}{!}{
\begin{tabular}{ll|S[table-format=2.3]S[table-format=2.3]S[table-format=2.3]S[table-format=2.3]S[table-format=2.3]S[table-format=2.3]S[table-format=2.3]}
\toprule
Dataset & $b_\text{sims}$ & \textsc{\small LLaMA2-7B} &  \textsc{\small LLaMA2-13B} &  \textsc{\small LLaMA2-70B} &  \textsc{\small MI$\odot$LLaMA2-7B} &  \textsc{\small MI$\odot$LLaMA2-13B} &  \textsc{\small MI$\odot$LLaMA2-70B} \\
\midrule
    Asia &   $10^0$ &     13.102 &      34.001 &      19.058 &        39.199 &         $\mathbf{59.664}$ &         49.702 \\
    &  $10^1$ &     44.274 &      37.710 &       8.753 &        46.244 &         $\mathbf{63.822}$ &         47.994 \\
     & $10^2$ &     40.333 &      25.254 &      13.228 &        $\mathbf{48.844}$ &         43.386 &         44.033 \\
\bottomrule
\end{tabular}

}
\end{center}
\end{table}

\section{Prompting Templates and Dataset Metadata}

The set of causal verbs under consideration~\citep{long2023causal} comprises 20 distinct wordings: \texttt{causes, provokes, triggers, leads to, induces, results in, brings about, yields, generates, initiates, produces, stimulates, instigates, fosters, engenders, promotes, catalyzes, gives rise to, spurs, sparks}.

Below, we give the prompt templates and dataset metadata that are used to compose the LLM queries. Note that they are provided herein for completeness and that they will be made available together with source code in a future version.

\begin{promptbox}{3-Way: Query}
\texttt{Among these three options which one is the most likely true: \\
(A) \$xvar \$verb \$yvar \\
(B) \$yvar \$verb \$xvar \\
(C) no causal relationship exists between \$xvar and \$yvar \\
The answer is:
}
\end{promptbox}

\begin{promptbox}{3-Way: Variable List}
\texttt{A list of all the phenomena to be considered follows. \\
\$varlist
}
\end{promptbox}

\begin{promptbox}{3-Way: Example}
\texttt{Among these three options which one is the most likely true: \\
(A) the presence of rain \$verb whether the grass is wet \\
(B) whether the grass is wet \$verb the presence of rain \\
(C) no causal relationship exists between the presence of rain and whether the grass is wet \\
The answer is: (A)
}
\end{promptbox}

\begin{promptbox}{3-Way: Priming}
\texttt{A User interacts with an Expert. \\
The Expert has profound knowledge and experience in \$domain. \\
The Expert answers queries about possible causal relationships between two phenomena using their knowledge about cause and effect in this domain. \\
The Expert answers either (A) or (B) corresponding to the direction of the causal relationship, or (C) if no causal relationship exists. \\
}
\end{promptbox}

\begin{promptbox}{2-Way: Query}
\texttt{Among these two options which one is the most likely true: \\
(A) \$xvar \$verb \$yvar \\
(B) \$yvar \$verb \$xvar \\
The answer is:
}
\end{promptbox}

\begin{promptbox}{2-Way: Variable List}
\texttt{A list of all the phenomena to be considered follows. \\
\$varlist
}
\end{promptbox}

\begin{promptbox}{2-Way: Example}
\texttt{Among these two options which one is the most likely true: \\
(A) the presence of rain \$verb whether the grass is wet \\
(B) whether the grass is wet \$verb presence of rain \\
The answer is: (A)
}
\end{promptbox}

\begin{promptbox}{2-Way: Priming}
\texttt{A User interacts with an Expert. \\
The Expert has profound knowledge and experience in \$domain. \\
The Expert answers queries about possible causal relationships between two phenomena using their knowledge about cause and effect in this domain. \\
The Expert answers either (A) or (B) corresponding to the direction of the causal relationship. \\
}

\end{promptbox}

\begin{promptbox}{Metadata: Asia}
\texttt{Domain description: pneumonology, a medical specialty that deals with diseases involving the respiratory tract \\
Variable id, name, and metadata: \\
1,asia,visited Asia \\
2,tub,tuberculosis \\
3,smoke,smoking cigarettes \\
4,lung,lung cancer \\
5,bronc,bronchitis \\
6,either,individual has either tuberculosis or lung cancer \\
7,xray,positive xray \\
8,dysp,"dyspnoae, laboured breathing "
}
\end{promptbox}

\begin{promptbox}{Metadata: Child}
\texttt{Domain description: neonatology, a subspecialty of pediatrics that consists of the medical care of newborn infants \\
Variable id, name, and metadata: \\
1,BirthAsphyxia,lack of oxygen to the blood during the infant's birth \\
2,Disease,infant methemoglobinemia \\
3,Age,age of infant at disease presentation \\
4,LVH,thickening of the left ventricle \\
5,DuctFlow,blood flow across the ductus arteriosus \\
6,CardiacMixing,mixing of oxygenated and deoxygenated blood \\
7,LungParench,the state of the blood vessels in the lungs \\
8,LungFlow,low blood flow in the lungs \\
9,Sick,presence of an illness \\
10,HypDistrib,low oxygen areas equally distributed around the body \\
11,HypoxiaInO2,hypoxia when breathing oxygen
12,CO2,level of CO2 in the body \\
13,ChestXray,having a chest x-ray \\
14,Grunting,grunting in infants \\
15,LVHreport,report of having LVH \\
16,LowerBodyO2,level of oxygen in the lower body \\
17,RUQO2,level of oxygen in the right up quadricep muscule \\
18,CO2Report,a document reporting high level of CO2 levels in blood \\
19,XrayReport,lung excessively filled with blood \\
20,GruntingReport,report of infant grunting 
}
\end{promptbox}

\begin{promptbox}{Metadata: Insurance}
\texttt{Domain description: car insurance, which provides financial protection against physical damage, injury, and liability resulting from traffic collisions, theft, or natural disasters \\
Variable id, name, and metadata: \\
1,Age,age \\
2,SocioEcon,socioeconomic status \\
3,RiskAversion,being risk averse \\
4,GoodStudent,being a good student driver \\
5,SeniorTrain,received additional driving training  \\
6,DrivingSkill,driving skill \\
7,MedCost,cost of medical treatment \\
8,OtherCar,being involved with other cars in the accident \\
9,MakeModel,owning a sport car  \\
10,VehicleYear,year of vehicle \\
11,HomeBase,neighbourhood type \\
12,AntiTheft,car has anti-theft \\
13,DrivHist,driving history \\
14,DrivQuality,driving quality \\
15,Airbag,airbag \\
16,Antilock,anti-lock \\
17,RuggedAuto,ruggedness of the car \\
18,CarValue,value of the car \\
19,Mileage,how much mileage is on the car \\
20,Accident,severity of the accident \\
21,Cushioning,quality of cushioning in car \\
22,Theft,theft occurred on the car \\
23,ILiCost,inspection cost \\
24,OtherCarCost,cost of the other cars \\
25,ThisCarDam,damage to the car \\
26,ThisCarCost,costs for the insured car \\
27,PropCost,ratio of the cost for the two cars 
}
\end{promptbox}

\end{document}